# Machine Vision-Based Crop-Load Estimation Using YOLOv8

Dawood Ahmed, Ranjan Sapkota, Martin Churuvija, and Manoj Karkee[*]

*Center for Precision & Automated Agricultural Systems, Washington State University, 24106 N Bunn Rd, Prosser, 99350, Washington, USA*

**Abstract**

*Shortage of labor in fruit crop production has become a significant challenge in recent years. Therefore, mechanized and automated machines have emerged as promising alternatives to labor-intensive orchard operations such as harvesting, pruning, and thinning. The use of mechanized and automated machines in fruit crop production has become a promising solution to the shortage of labor, as these technologies can accomplish labor-intensive tasks such as harvesting, pruning, and thinning. One of the key aspects of agricultural robots in accomplishing these tasks is their ability to identify tree canopy parts such as trunks and branches and estimating their geometric and topological parameters such as branch diameter, branch length, and branch angles. Having an estimate of the target crop-load, researchers then can work on automated pruning and thinning platforms that make more effective decisions to achieve optimal crop yields. In this study, we propose a machine vision system to estimate these canopy parameters in apple orchards. These parameters were then used to estimate an optimal number of fruit that individual branches could bear in a commercial orchard, which provides a basis for robotic pruning, flower thinning, and fruitlet thinning so that desired fruit yield and quality could be achieved. Utilizing color and depth information collected with an RGB-D sensor (Microsoft Azure Kinect DK, a YOLOv8-based instance segmentation technique was developed to identify trunks and branches of apple trees in the dormant season. We then applied a Principal Component Analysis technique to estimate branch diameter (used to calculate limb cross-sectional area or LCSA) and orientation. The estimated branch diameter was used to calculate the limb cross-sectional area (LCSA), which was then used as an input for crop-load estimation, as a larger LCSA indicates a higher potential fruit-bearing capacity of the branch. RMSE for branch diameter estimation was calculated to be 2.08 mm and for crop-load estimation to be 3.95. Based on the established management practices in commercial apple orchards, we estimated the target crop-load (number of fruit) for each segmented branch with a mean absolute error (MAE) of 2.99 and (ground truth crop-load was 6 apples per LCSA). Our study demonstrated a promising workflow with a high level of performance in identifying trunks and branches of apple trees in a dynamic commercial orchard environment and integrating farm management practices into automated decision-making.*

*Keywords:* Object detection,YOLOv8, deep learning, machine vision, agricultural automation and robotics

## 1. Introduction

Around 30-40 % of the total value of the United States (U.S.) crops belong to specialty crops (Fuchs et al., 2021), which clearly shows the importance of this industry to U.S. agriculture and economic activities.  More than 200 thousand seasonal workers are invited each year from other countries to work in the tree fruit orchards in U.S. performing field operations such as tree pruning, flower thinning, green fruit thinning, and harvesting. However, in recent decades, growers are facing challenges in finding enough farm labor to complete field operations in various specialty crops (Bogue, 2020). Recently, the farm labor crisis has become worse because of the global pandemic as the reduction of agricultural labor inputs due to the COVID-19 pandemic has resulted in an estimated loss of $309 million in agricultural production over the first year of Pandemic (March, 2020, to March, 2021; Bochtis et al., 2020; Lusk & Chandra, 2021). Thus, the agricultural production system is in critical need of automated and robotic machines that could operate in orchard environments to perform the various crop- and canopy-management operations (Sapkota et al., 2023; Q. Zhang et al., 2019)

Robots, with a robust machine vision and manipulation systems, have the potential to reduce human labor by making decisions and acting in real-time to perform specific or programmed tasks requiring repeatable accuracy(Bechar, 2021). Researchers and engineers, since the last two decades, have been investigating the development of mechanized and automated platforms that could mimic human operation in the orchard environment such as automated fruit picking (Hua et al., 2019; Huang et al., 2020; Kondo et al., 1996; Verbiest et al., 2021), automated tree pruning(He & Schupp, 2018; Liu et al., 2012), robotic shoot thinning (Majeed et al., 2020, 2021) and automated flower thinning (Nielsen et al., 2011; C. Zhang et al., 2022).

Despite the tremendous research and development efforts in recent years, there has been no reported automated machines commercially adopted so far for crop-load management operations such as selective tree pruning and fruit thinning in the real-world environment. Due to the complex, unstructured/uncontrolled environment, and unpredictable variability in lighting, landscape, and atmospheric conditions, the demands of motions of robots for the automated crop-load management operation changes often in time and space, making it an even more complex problem (Bechar & Vigneault, 2016). Crop-load management in U.S. orchards follows specific guidelines, such as tree pruning, flower thinning, and fruit picking thinning (Q. Zhang et al., 2019). Nonetheless, efficiently managing the desired amount of crop-load through an automated decision system remains a significant challenge for researchers and engineers. They are currently working on developing a precise guidance system for robots to manage the crop-load according to these guidelines.

Fruit tree pruning involves selectively cutting and removal of certain branches of a tree following some guidelines (provided by researchers or experienced growers) that allow the fruits to grow uniformly at desired locations. Additionally, tree pruning ensures proper penetration of sunlight and air into the canopies and regulates enough fruiting site space to achieve better yield and crop quality. Likewise, the thinning operation involves removal of a portion of flower bloom and/or immature fruits (fruitlets) to ensure desired quality (including size) of the remaining fruit. These crop-load management operations in tree fruit production is a balancing act between maximizing yield (crop-load) while optimizing fruit quality and ensuring adequate return bloom (Robinson et al., 2014). Thus, to effectively perform most of these crop-load management operations using a robotic system, it is critical to estimate the target crop-load in each of these branches, which is determined, in commercial apple farming, based on tree traits information such as trunk size, branch diameter, and branch length. The branches support the growth of leaves, fruits, flowers and buds, whose geometric properties can provide insights about normal growth, fruiting and flowering information because they are the substantial indicator of crop growth and yield (Nyambati & Kioko, 2018). Therefore, automated estimation of geometric parameters/traits (e.g. length and diameter) of fruit tree branches could provide crucial information to further advance robotic systems for various crop-load management operations.

Over the past decade, various 3D reconstruction and branch recognition techniques have been proposed and tested in commercial orchards. Karkee et al. (2014) used a time-of-flight-based 3D camera for the 3D reconstruction of apple trees and showed how such a geometrical representation could be useful in developing pruning rules. Tabb et al. (2017) used a robot vision system (Robotic System for Tree Shape Estimation (RoTSE)) to reconstruct apple trees in the dormant season using a shape-from-silhouette method and use the structure to measure geometric attributes of apple trees such as branch structure, diameters, lengths, and branch angles. Likewise, Tong et al. (2022), and Zhang et al. (2017 and 2020) are some recent studies in detecting branches in apple trees using deep learning techniques such as Regions-Convolutional Neural Network (R-CNN), Faster R-CNN, Mask R-CNN, and Cascade Mask R-CNN respectively. Song et al. (2021) have recently developed a handheld device for measuring the diameter at breast height (DBH) of a tree using a digital camera, laser ranging, and image recognition. The handheld device was designed to perform instant, automated, and non-contact DBH measurements. However, it had a measurement bias up to -1.78 mm. Likewise, Yuan et al. (2021) developed an Intelligent Electronic Device (IED) to measure DBH and tree height through high-precision hall angle sensors and a dual in-line package (DIP) sensor. However, the author in this study reported an estimation bias of 1.6 mm compared with the caliper reading. Similarly, to measure tree DBH remotely, Fan et al. used RGB-D images collected using a mobile phone and simultaneous localization and mapping (SLAM). However, the author reported a large measurement bias in this case as well while validating the study (Fan et al., 2018). Likewise, in order to perform outdoor measurement of tree stem DBH, McGlade et al. recently made used a RGB-D sensor called Kinect V2 where the author recorded 51 individual urban trees from one viewing angle at a distance of 1 m to 5 m away using various Field of View (FOV) settings on the depth sensor (McGlade et al., 2020). The author then implemented a circle-fitting approach on resultant point clouds to estimate the DBH. The study provided higher RMSE error compared to the ground truth of stem and was only feasible for those trees where non-circular and irregular stems were removed.

Despite all these studies, extracting accurate 3D attributes of tree structures in unstructured and dynamic orchard environments remains challenging due to numerous obstacles such as occlusions, lighting variations, background noise, and inherent limitations of sensors (Akbar et al., 2016). Additionally, the existing studies regarding the tree traits identification (e.g. branch and trunk detection) could not provide any insights on how to use that detected branch/trunk information through an automated platform for crop-load management in commercial agriculture. Although studies have been reported on estimating crop-load in apple orchards using machine vision-based apple detection and counting (Aggelopoulou et al., 2011; Gongal et al., 2016; Ji et al., 2012; Linker et al., 2012), this approach does not satisfy the need of estimating target crop-load during the early growing season because the fruiting information cannot increase the fruits if the orchard is under cropped. Thus, to facilitate the development and enable implementation of various crop-load management strategies and rules (e.g. pruning and thinning), it is necessary to acquire information related to the observable characteristics of the trees (such as their physical traits and features), which would facilitate the development and implementation of crop-load management strategies and rules, such as pruning and thinning.

Regardless of numerous research efforts, the existing studies have been able to propose only a theoretical overview of estimating tree trunk and branch diameters, without practical application in commercial farming environments. Furthermore, the utilization of branch size information for crop-load management in real-world orchard settings has been largely unaddressed. Consequently, the primary objective of this study is to develop a robot vision system capable of estimating the target crop-load in a commercial apple orchard during the dormant season. This estimation will not only provide insights into the potential fruit-bearing capacity of individual branches but also assist in making informed automated pruning decisions, as farmers typically prune commercial trees during this period. By bridging the gap between theoretical understanding and practical application, this research aims to contribute to the development of efficient and accurate crop-load management strategies, enabling improved decision-making in commercial apple orchards. The following materials and methods section will detail the approach and techniques employed to achieve these objectives.

## 2. Materials and Methods

This study can be divided into four steps; i) acquire image data; ii) detect and segment the branches and trunks of apple trees using a deep learning approach; iii) Extract the 3D point clouds from the segmented branch mask and estimate the limb cross-section area (LCSA); iv) Estimate the desired crop-load in number on a particular branch section by integrating a commercial management approach. Figure 1 shows the functional workflow diagram of this study.

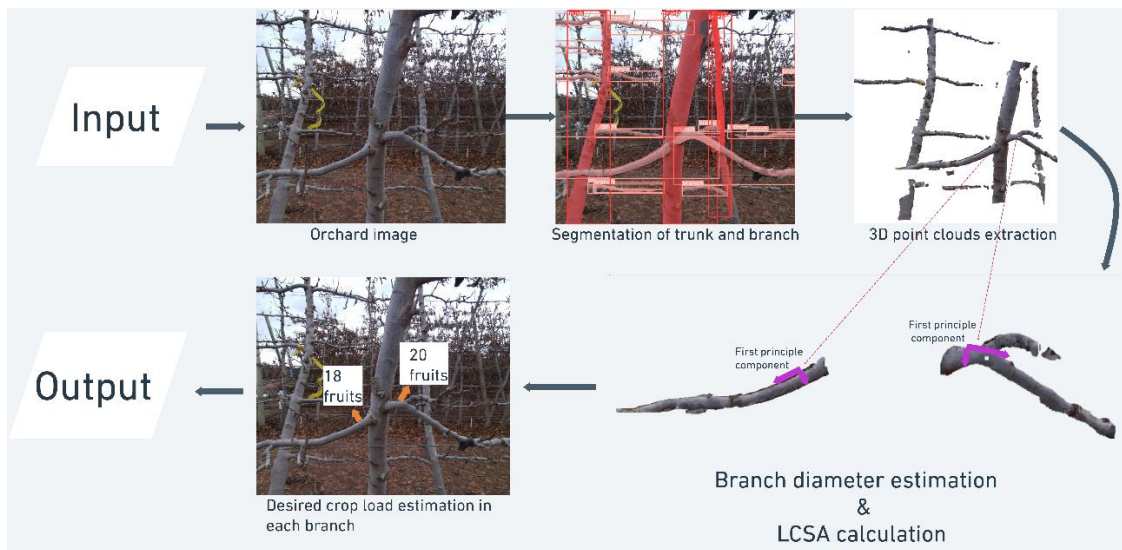

*Figure 1: Workflow diagram of the study on machine vision system for precise crop load estimation in commercial orchard using deep learning method and branch diameter property*

## 2.1 Study site and data acquisition

This study was carried out in a commercial apple orchard (Allan Brothers Orchard, Figure 2a) located in Prosser, Washington, United States. The orchard was planted with an Envy variety trained to a V-trellis architecture. Trees were planted in 2009 with a row spacing of 9.0 ft and a plant spacing of 3.0 ft. A set of RGB-D images were acquired using an Intel RealSense 435i camera (Figure 2b) (Intel RealSense Technology, California, USA) and a Microsoft Azure Kinect DK AI camera (Figure 2c) (Microsoft Azure, Redmond, Washington) in December 2021 and November 2022 respectively. RGB images from the Intel RealSense camera were used to implement the deep learning method to identify the branches and trunk of apple trees whereas RGB-D images from Microsoft Azure camera were used to estimate the diameter and LCSA of the branch for target crop-load estimation.

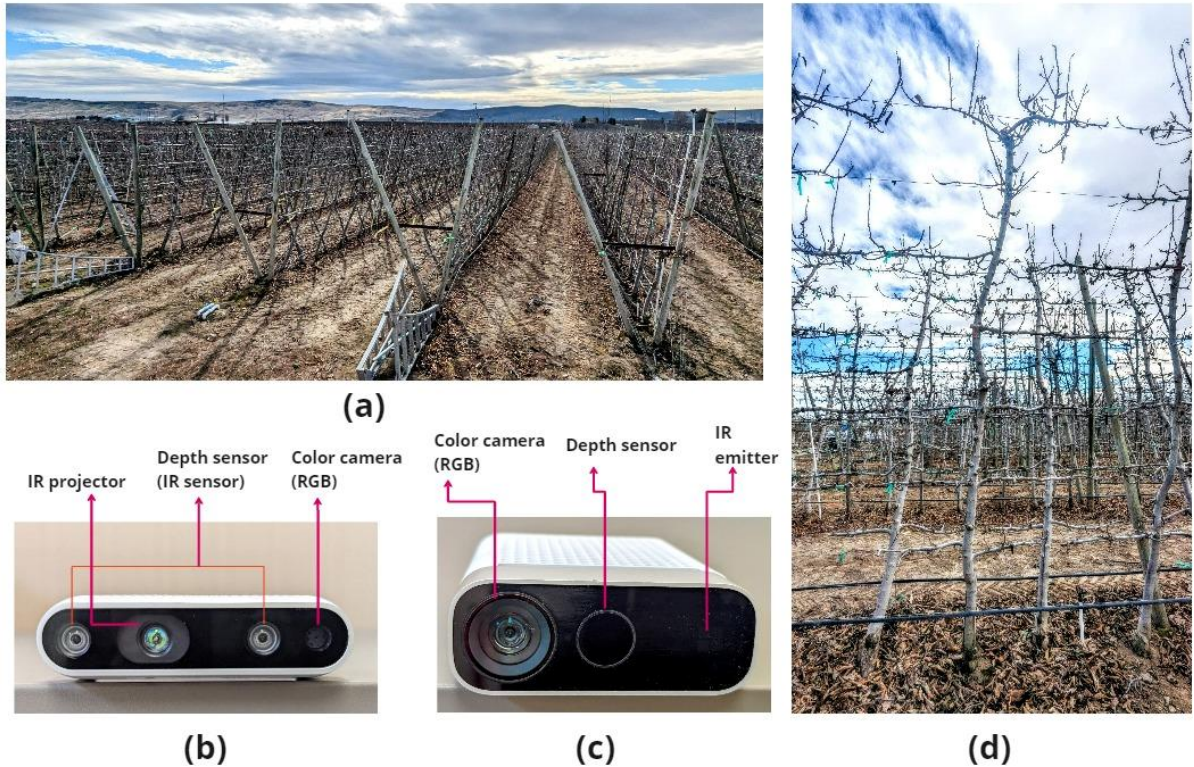

*Figure 2: (a) Commercial apple orchard used in this study (Prosser, Washington). (b) Intel Realsense D435i RGB-D camera, (c) Microsoft Azure Kinect DK camera; and (d) Example image of apple trees used in the study.*

Structured light (SL) and stereoscopy (SC) techniques were by the Intel RealsenseD435i camera to estimate 3D information of the scene whereas Microsoft Azure AI camera was based on principle of Time-of-Flight (ToF) of light. Table 1 presents the additional specifications of the vision sensors used in this study.

*Table 1: Specification of the sensors/cameras used in this study (SL: Structured light; SC: Stereo Camera)*

| Camera | Principle | Measuring Range(m) | Depth Resolution | RGB Max Resolution | Frame rate | FoV Depth | Price |
|---|---|---|---|---|---|---|---|
| Realsense D435 | SL+SC | 0.11-10 | 1280 x 720 | 1920 x 1080 | 90 | 85.2° × 58° | 314 |
| Microsoft Azure | ToF | 0.5-5.4 | 640 x 576 | 4000 x 3000 | 30 | 120° × 120° | 399 |

Cameras were positioned at 1 m (approximate) distance from the tree trunk, and images were captures from various camera heights above the ground including sunny and cloudy conditions.

## 2.2 Detection and segmentation of trunk and branch using YOLOv8 model

YOLO (You Only Look Once) is a widely used object detection framework that is known for its high accuracy and performance (Redmon et al., 2016a). YOLO was first introduced by Redmon et al. (2016)on the paper "You Only Look Once: Unified, Real-Time Object Detection", which is called a single-stage detector because it does everything in one step (Jiang et al., 2022). YOLOv8 is the latest version of this framework which was created in January, 2023 by Ultralytics (Ultralytics, Maryland, USA), works by dividing an image into a grid of smaller regions and then predicting a bounding box and class probabilities for each object that is present in each region. The YOLOv8 algorithm uses the Darknet-53 architecture to improve the feature extraction process, leading to more accurate object detection. DarkNet-53 is a convolutional neural network with 53 layers and can classify images into 1,000 object categories. This network is divided into smaller stages, and then connects these stages in a partial way, allowing for better feature reuse and gradient propagation.

One of the key improvements in YOLOv8 over previous versions is that it incorporates a technique called PS (Pseudo Ensemble or Pseudo Supervision), which involves the use of multiple models with different configurations during the training process. These models are trained on the same dataset, but with different hyperparameters, leading to a more diverse set of predictions. During inference, the predictions from different models are combined to produce a final prediction, leading to better accuracy and robustness. This technique is particularly useful when there is a limited amount of annotated training data available, as it allows the model to learn from its own predictions and generate a more diverse and accurate output. Figure 3 is the architecture of YOLOv8 algorithm implemented in this study for the detection of branch and trunk in apple tree images.

Earlier version, YOLOv5, included an efficient backbone network, a higher-resolution feature pyramid, and improved anchor boxes that adapt to different object shapes and sizes(Ge et al., 2021). YOLOv5 also incorporated advanced training techniques such as cutcut mixed mosaic augmentation, self-adversarial training, and a focal loss function to further improve its performance (X. Zhang et al., 2021). At its core, YOLOv5 works by first dividing the input image into a grid of cells and predicting the likelihood of an object being present within each cell (Ge et al., 2021). Each grid cell predicts a fixed number of bounding boxes, along with the class probabilities for each box. The network then refines the bounding box predictions based on the contents of the cell and the surrounding cells in the feature map. This allows YOLO models to accurately detect objects at different scales and locations within the image while in contrast to other YOLOs, YOLOv8 combines the architecture of YOLOv4, DarkNet-53, and PS to improve the accuracy and robustness of object detection. There are five different versions of YOLOv8 as YOLOv8n-seg, YOLOv8s-seg, YOLOv8m-seg, YOLOv8l-seg and YOLOv8x-seg.

In this study, we employed the YOLOv8-based deep learning approach (Figure 3) to detect and segment the trunks and branches of apple trees using RGB images from the Intel RealSense 435i camera. The input data consisted of 474 annotated images, partitioned into training, validation, and test sets in an 8:1:1 ratio. There were1,141 labels for the tree trunk and 2,369 labels for the tree branches connected to the trunk created in the format of the COCO dataset generated using Labelbox (Labelbox, San Fransisco, US).

The YOLOv8 model was chosen due to its ability to efficiently detect objects while maintaining high accuracy. The model's architecture (Figure 4) was configured with DarkNet-53 for improved feature extraction and the Pseudo Ensemble (PS) technique for enhanced robustness in predictions. The model's output included bounding boxes and class probabilities for each detected trunk and branch, providing critical information for further analysis. The selected parameters, such as learning rate, batch size, and optimizer, were carefully chosen through multiple training and debugging trials to optimize the model's performance. The final hyperparameters were set to ensure a balance between model accuracy and computational efficiency. For instance, a batch size of 16 allowed for faster training while maintaining model stability, and an initial learning rate of 0.01 facilitated effective weight updates. Momentum and weight decay parameters were also included to accelerate convergence and prevent overfitting, respectively. The deep learning framework was implemented using PyTorch, and the YOLOv8 model was trained on a Linux System with an eight-core Intel i7 CPU and an RTX 3070 graphics card. By employing this state-of-the-art object detection method, we were able to accurately identify and segment trunks and branches of apple trees, which served as a crucial step towards estimating limb cross-section area and determining the desired crop-load on individual branch sections. Additional information about metrics applied to image during model training is presented in Table 2.

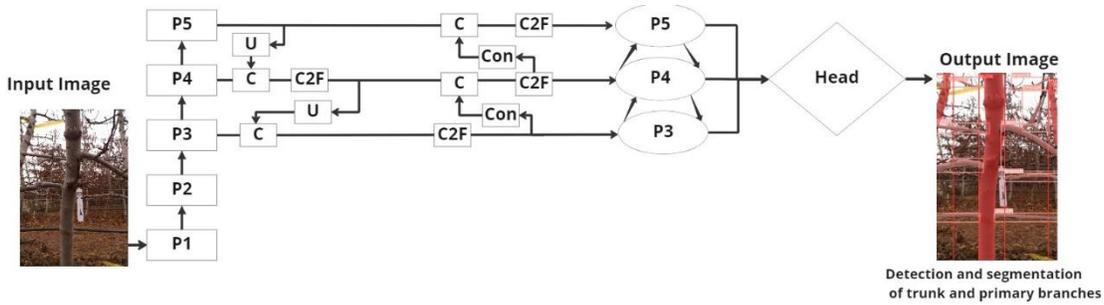

*Figure 3: Processing image through YOLOv8 object detection model to detect the trunk and branch. The diagram shows the surface architecture of YOLOv8*

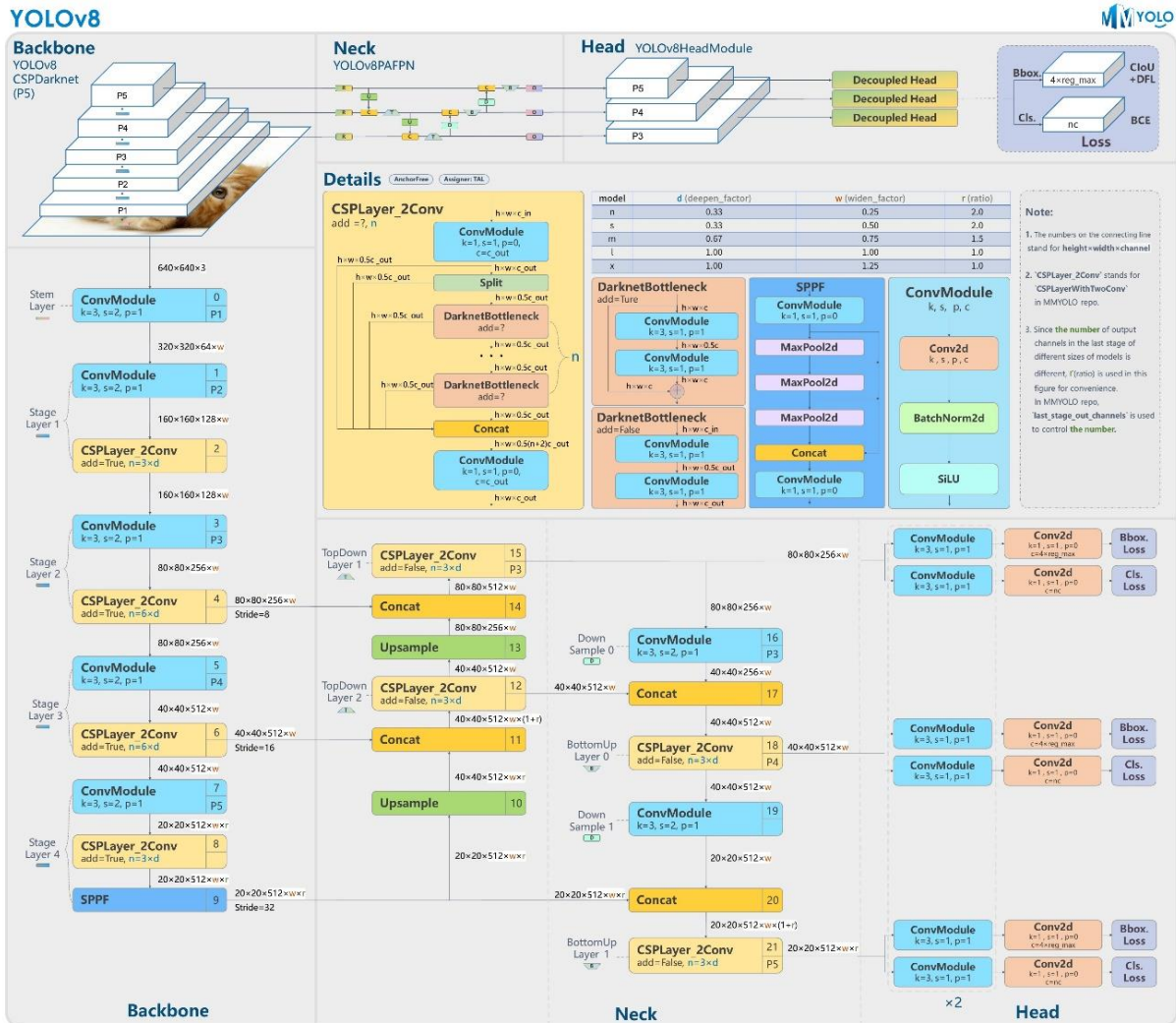

*Figure 4: Functional diagram of YOLOv8 (Ultralytics (Version 8.0.0) [Computer Software]. Https://Github.Com/Ultralytics/Ultralytics)*

*Table 2: Metrics and processes applied to the images*

| Methods Applied | Value |
|---|---|
| Hue augmentation (fraction) | **0.015** |
| Saturation augmentation (fraction) | 0.7 |
| Value augmentation (fraction) | 0.4 |
| Rotation | 0.0 |
| Translation | 0.1 |
| Scale | 0.5 |
| Flip left-right (probability) | 0.5 |
| Mosaic (probability) | 1.0 |
| Weight decay | 0.0005 |

## 2.3 Principal Component Analysis for 3D branch orientation

Principal Component Analysis (PCA) is a widely utilized unsupervised machine learning technique that identifies patterns and relationships in data without relying on pre-existing labels or guidance. By reducing data dimensionality, PCA enhances interpretability while minimizing information loss (Abdi & Williams, 2010). It is instrumental in identifying the most significant features in specific datasets. In this study, PCA was employed to determine the principal components with the highest variation in the 3D point clouds extracted from the branch masks detected by the YOLOv8 model in RGB images. The 3D point clouds extracted from the branch masks were represented by the direction and magnitude of variation. For this analysis, data normalization was necessary, as unscaled data with different measurement units could distort the comparison of the magnitude of variance among various features. Data normalization was achieved by subtracting the mean and dividing it by the standard deviation for each variable, transforming all variables to the same scale, explained as equation 1.

$$Z = \frac{x - \mu}{\sigma}$$

*Equation 1*

Following data standardization, the covariance matrix was computed. This computation estimated how data deviated from the mean across multiple dimensions. In this study, three dimensions of point clouds were considered as three unique characteristics, and a 3x3 covariance matrix was constructed to analyze the correlation between different dimensions. The diagonal elements of the matrix reflected the variance of each feature, while the non-diagonal elements represented the variance between two distinct features. This information was vital for feature set reduction, as it enabled the detection of redundant features and the assessment of the cumulative proportion of variance contributed by each feature.

$X$ = Point values is X-axis
$Y$ = Point values in Y-axis
$Z$ = Point values in Z-axis

$$\text{Covariance} = \begin{matrix} cov(X,X) & cov(X,Y) & cov(X,Z) \\ cov(Y,X) & cov(Y,Y) & cov(Y,Z) \\ cov(Z,X) & cov(Z,Y) & cov(Z,Z) \end{matrix}$$

*Equation 2*

To identify the principal components, the covariance matrix was first computed, followed by the calculation of eigenvalues and eigenvectors. Principal components are uncorrelated new variables that account for varying proportions of the variance. Three principal components were obtained from the three-dimensional data, with each principal component having one eigenvector representing direction and one eigenvalue representing variance magnitude. In this study, the 3D point clouds extracted from the branch mask had one eigenvector representing the direction of the 3D point clouds and another eigenvalue representing the diameter of the branch at that point. The second eigenvalue was obtained by drawing a perpendicular line from the branch's boundary. Since only the most significant direction vector for each branch was required, only the first principal component was utilized in this study.

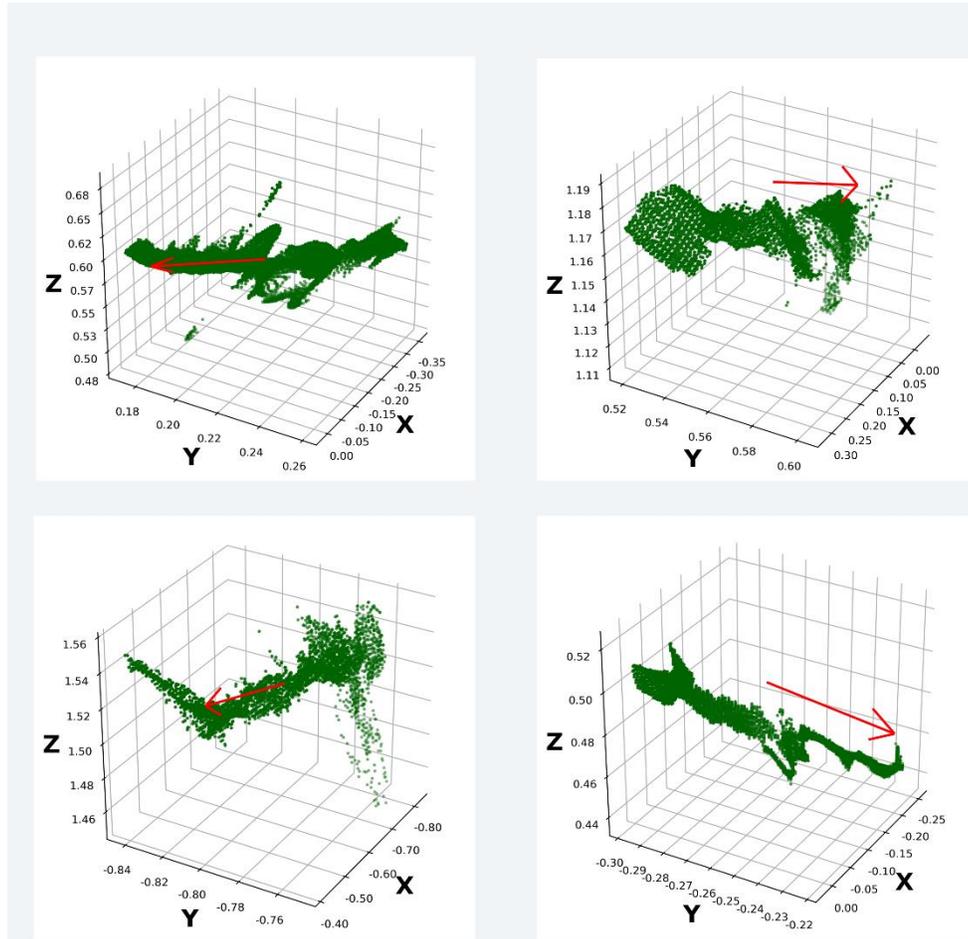

*Figure 5: Principal Component Analysis samples on scattered 3D point cloud data of branch section samples segmented by YOLOv8 object detection and segmentation model.*

## 2.4 Branch Diameter Estimation

The point cloud of a branch segment, extracted from the branch mask identified by the YOLOv8 model and processed through PCA, was utilized to estimate the branch diameter in 3D space. To determine the appropriate location for diameter measurement on the branches, a comprehensive review of the literature and consultations with multiple commercial apple growers were conducted. It was found that the common practice in commercial orchards involves measuring the diameter at a region approximately 3 cm away from the trunk-branch boundary. This approach is based on the understanding that taking measurements less than 3 cm away from the trunk boundary yields more accurate results for the associated branch. In accordance with this guideline, branch diameters (ground truth data) were collected, as demonstrated in Figure 6.

For the 3D point clouds, as demonstrated in figure 5, the point of measurement is the normal plane cutting the branch points extremely close to the trunk end of the branch. We locate the normal plane of the branch and estimate its diameter by measuring its length along the normal axis using the orientation information retrieved using PCA. This method works well if the depth camera produces an accurate depth profile of the branch. It also relies on the instance segmentation model to capture sufficient segmentation information.

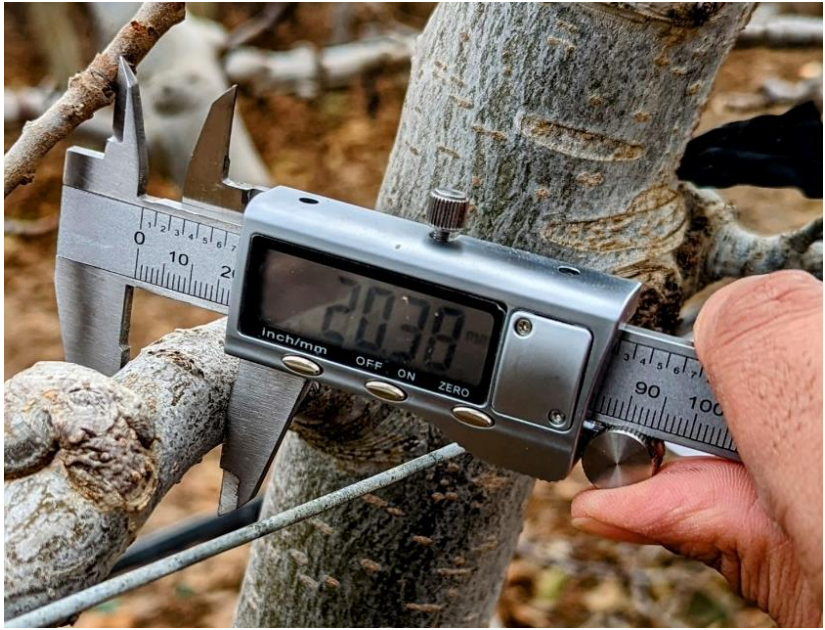

*Figure 6: Ground truth collection for diameter of primary branch using digital calliper*

## 2.5 Crop-Load Estimation

According to some recently reported studies, for the vast majority of apples, keeping 6 fruits per cm square limb cross-sectional area  LCSA have been found to be best in terms of improved fruit set, fruit weight, mineral composition and return bloom. Sidhu et al have recently compared the efficacy of thinning based on artificial bud extinction (ABE) by keeping 3, 6 and 12 fruits per cm square LCSA on 'scilate' apples and found 6 fruit per cm square to be the optimal approach(Sidhu et al., 2022a) . Likewise, Anthony et al. have performed a comparative study on 'W28' apples (which is a combination of 'Enterprise' and 'Honeycrisp' apples) by adjusting 2,4,6, and 8 fruits per $cm^2$ trunk cross sectional area (TCSA) and concluded that the optimal fruit quality and bloom was identified at 6 fruits per $cm^2$ TCSA (Anthony et al., 2019).

Additionally, to estimate the desired crop-load for each branch, we took insights from the farm manager and growers of the Olsen Bros Ranches, Inc. which is a fruit-producing and wholesale company located in Prosser. The growers described the "6 apples per cm² (LCSA)" concept used in apple farming as a way to estimate the number of apples a particular limb on a tree can support while maintaining the overall health and productivity of the tree. The approach is based on the idea that each limb of an apple tree can only support a certain amount of fruit based on its cross-sectional area. The guideline suggests that each square centimeter of a limb's cross-sectional area can support up to six apples, i.e if a particular limb has a cross-sectional area of X square centimeters, it can support up to 6X apples.

$$\textit{Desired Crop Load} = 6 \; x \; \textit{LCSA}$$
*Equation 3*

The LCSA of the branch section was calculated by using the formula of area circle which is given as:

$$LCSA = \frac{\pi d^2}{4}$$
*Equation 4*

Prior to adopting the "6 fruits per cm square LCSA" method for desired crop-load estimation in this study, a commercially available tool for crop load estimation was examined. This tool, as depicted in Figure 7, is utilized during manual pruning and thinning and offers guidance on the desired crop-load for each branch of an apple tree. The device, employed by Allan Bros Fruit Company in Washington state, measures the LCSA of each branch and provides target and desired crop-load values in numbers. This strategy, which suggests a specific number of fruits per cm², allows growers to estimate the number of apples that each limb can support, enabling them to thin the fruit accordingly to achieve the desired crop-load. It is important to note that, while commercial operations may be referenced for the adoption of certain strategies, scientific findings should not solely rely on growers' practices or experiences, unless supported by a well-designed scientific user study.

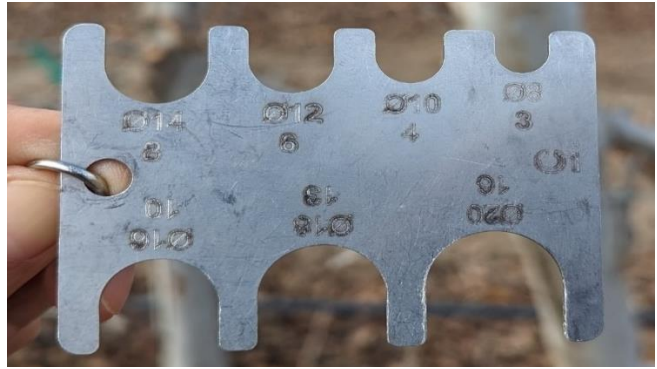

*Figure 7: Device being used by commercial growers to estimate desired crop load using branch LCSA property*

## 2.6 System Evaluation

The detection and segmentation performance of the YOLOv8 algorithm is assessed using three main evaluation indicators: Mean Intersection over Union (IoU), Average Precision (AP), and Mean Average Precision (mAP). The IoU, also referred to as the Jaccard index, quantifies the degree of overlap between the segmented mask and the target object, effectively gauging the accuracy of the segmentation process. On the other hand, AP measures the area enclosed by the recall rate, precision rate, and the horizontal axis, providing an evaluation of target detection performance. Lastly, the mAP serves as an aggregate metric, encompassing the performance of both target detection and instance segmentation, offering a comprehensive assessment of the YOLOv8 algorithm's efficacy. The calculation equations are as follows:

$$MIoU = \ = \frac{Area\ Overlap}{Area\ Union} = \frac{TP}{FP + TP + FN}$$

*Equation 5*

$$mAP = \frac{TP}{TP + FP}$$

*Equation 6*

$$mAR = \frac{TP}{TP + FN}$$

*Equation 7*

$$f1 - Score = \frac{2(Precision * Recall)}{Precision + Recall}$$

*Equation 8*

where TP, TN, FP and FN are true positive, true negative, false positive and false negative respectively.

For evaluation of the branch diameter estimation and crop-load estimation (desired), Root Mean Squared Error (RMSE) was considered which is given as:

$$RMSE = \sqrt{(\frac{1}{n}\sum(predicted\_i - actual\_i)\wedge 2)}$$

*Equation 9*

In this equation, 'n' represents the total number of samples, while 'predicted_i' and 'actual_i' denote the predicted and actual values, respectively, for each sample 'i'. The actual target in this case refers to the true value of branch diameter, and the desired crop-load estimation is derived from the branch diameter estimation. Consequently, the RMSE of the branch diameter estimation effectively captures the error associated with the number of fruits that can be supported by each branch. This measure provides a useful reference for growers, enabling them to understand the degree of error they may encounter when using this technique for crop-load estimation.

Additionally, the estimated branch diameter and crop-load were compared with the respective ground truth values using Mean Absolute Error (MAE) given by:

$$MAE = (\frac{1}{n}\sum|predicted_i - actual_i|)$$

*Equation 10*

In addition to the other evaluation metrics, the estimated branch diameter and crop-load were also assessed using the Mean Absolute Percentage Error (MAPE), which is an evaluation metric that quantifies the average percentage difference between predicted and actual values. The MAPE is calculated using the following formula:

$$MAPE = (\frac{1}{n}\sum \left|\frac{predicted_i - actual_i}{actual_i}\right| * 100$$

*Equation 11*

## 3. Results and Discussion

### 3.1 YOLOv8 performance evaluation

Figure 8 presents representative examples of the tree trunk and branch detection and segmentation achieved using the YOLOv8-based model. These results demonstrate the effectiveness of the model in accurately identifying and delineating the tree structures within the images. The visual outputs of the model not only highlight the potential of this approach in the context of the study but also offer insight into its practical application for tree analysis in orchard settings. Table 4 and 5 show the precision, recall and F1-score achieved by the model in segmenting tree trunks and branches respectively. Likewise, overall performances for both trunk and branches are presented in Table 6.

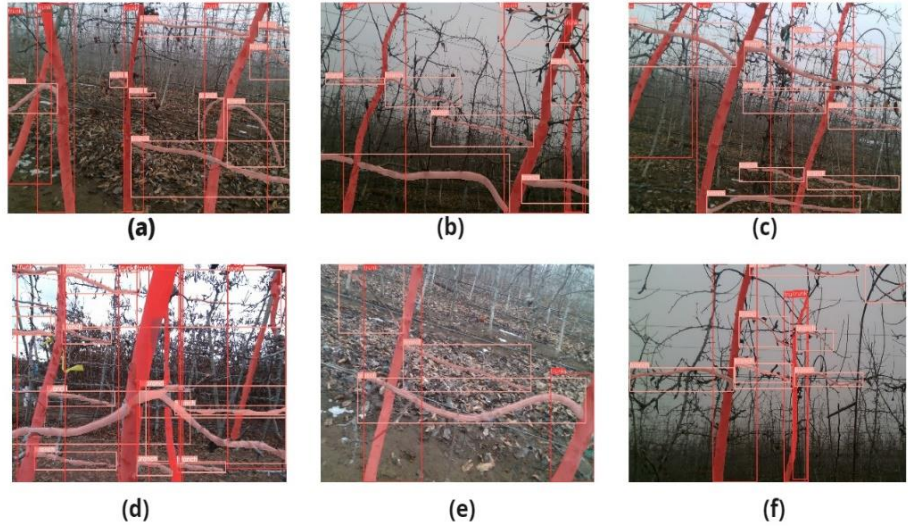

*Figure 8: Sample images showing the results of YOLOv8-based object detection and segmentation technique; Tree trunks and branches were segmented out in the images acquired in a commercial orchard during the dormant season. (a)low light condition, image depicting orchard in the background, (b, c) low light with foggy weather light in the background, (d) brighter light in the background (e) detection in low branch density areas (f) detection in low light and complex background*

*Table 2: Results and evaluation metrics for trunk segmentation*

| Model | Precision | Recall | F1-Score | AP@0.5 | AP@0.5:0.95 |
|---|---|---|---|---|---|
| YOLOv8n-seg | 0.86 | 0.93 | 0.89 | 0.96 | 0.58 |
| YOLOv8s-seg | 0.88 | 0.95 | **0.92** | 0.94 | 0.59 |
| YOLOv8m-seg | 0.86 | **0.97** | 0.91 | 0.95 | 0.63 |
| YOLOv8l-seg | 0.80 | **0.97** | 0.88 | 0.96 | 0.61 |
| YOLOv8x-seg | **0.88** | 0.95 | **0.92** | **0.97** | **0.63** |

*Table 3: Results and evaluation metrics for branch segmentation*

| Model | Precision | Recall | mAP@0.5 | mAP@0.5:0.95 |
|---|---|---|---|---|
| YOLOv8n-seg | 0.79 | 0.82 | 0.83 | 0.42 |
| YOLOv8s-seg | 0.80 | 0.84 | 0.84 | 0.43 |
| YOLOv8m-seg | 0.78 | **0.88** | 0.85 | **0.46** |
| YOLOv8l-seg | 0.77 | 0.83 | **0.85** | 0.45 |
| YOLOv8x-seg | **0.81** | 0.83 | 0.84 | 0.45 |

*Table 4: YOLOv8 model overall performances for both object class, Trunk and Branch*

| Model | Precision | Recall | F1 Score | AP@0.5 | AP@0.5:0.95 |
|---|---|---|---|---|---|
| YOLOv8n-seg | 0.72 | 0.72 | 0.72 | 0.70 | 0.26 |
| YOLOv8s-seg | 0.71 | 0.72 | 0.72 | 0.73 | 0.28 |
| YOLOv8m-seg | 0.70 | **0.79** | **0.74** | 0.74 | **0.30** |
| YOLOv8l-seg | **0.73** | 0.70 | 0.71 | **0.75** | 0.29 |
| YOLOv8x-seg | **0.73** | 0.70 | 0.71 | 0.71 | 0.27 |

The table displays the quantitative results for precision, recall, F1 score, and AP@0.5 values for the five YOLOv8 models in segmenting trunks and branches. For trunk detection, YOLOv8x-seg achieved the highest precision (0.88) and recall (0.95), while YOLOv8l-seg demonstrated the lowest precision (0.80). In terms of branch detection, YOLOv8x-seg displayed the highest precision (0.81) and recall (0.83), whereas YOLOv8m-seg and YOLOv8s-seg had slightly lower precision values (0.78 and 0.80, respectively). Based on these findings, YOLOv8x-seg was identified as the optimal model for both trunk and branch detection, exhibiting superior performance across all evaluation metrics. Consequently, the YOLOv8x-seg model was selected for further analysis in estimating branch diameter and crop-load, allowing for a consistent and focused evaluation of its practical application in achieving the study's objectives. Moving forward, the remaining analysis will be conducted using the best/optimal model selected in this early stage of the results and discussion.

The goal of this study is not merely to evaluate the YOLOv8 models but to identify the best model for addressing the problem at hand, which is estimating branch diameter. The various YOLOv8 models possess different characteristics, as described in the methods section, which result in distinct strengths and weaknesses for object segmentation tasks. Therefore, it is crucial to evaluate their performance based on various tasks and metrics. From the results, it is evident that YOLOv8m-seg and YOLOv8l-seg models are highly accurate in identifying trunk-related objects, while YOLOv8m-seg demonstrates superior performance for branch-related objects.

The F1-score offers a balanced measure between precision and recall, indicating that the YOLOv8x-seg model demonstrated a favorable equilibrium for detecting tree trunks, while the YOLOv8m-seg model exhibited a desirable balance for detecting tree branches. These insights can be instrumental in choosing the most suitable model for the specific task of detecting tree trunks or branches.

The results suggested that the model correctly identified 95% of instances for both trunk and branch classes, although some false positives were generated. The mean average precision (mAP) values of 0.95 and 0.74 in Figure 9 (c) corresponded to the trunk and branch class detection, respectively, at a threshold of 0.5. The overall mAP of 0.85 for all classes combined at the same threshold indicated the model's overall performance in object detection. In the F1-confidence curve (Figure 9d), an F1-score of 0.83 at a confidence threshold of 0.58 demonstrated the trade-off between precision and recall for both trunk and branch classes at that specific threshold. A higher F1-score signified better model performance in terms of precision and recall. These findings provided insights into the model's performance and could be used to assess its suitability for detecting and segmenting trunks and branches in orchard images.

The performance of the YOLOv8 family of models for object detection and segmentation in an apple orchard setting has been evaluated using the relationship between mAP@0.5 and FPS (Figure 10). The results show that YOLOv8x-seg model has the highest mAP of 0.88 for branch and 0.81 for trunk, while YOLOv8s-seg has the lowest mAP of 0.80 for trunk and 0.71 for branch among the models.

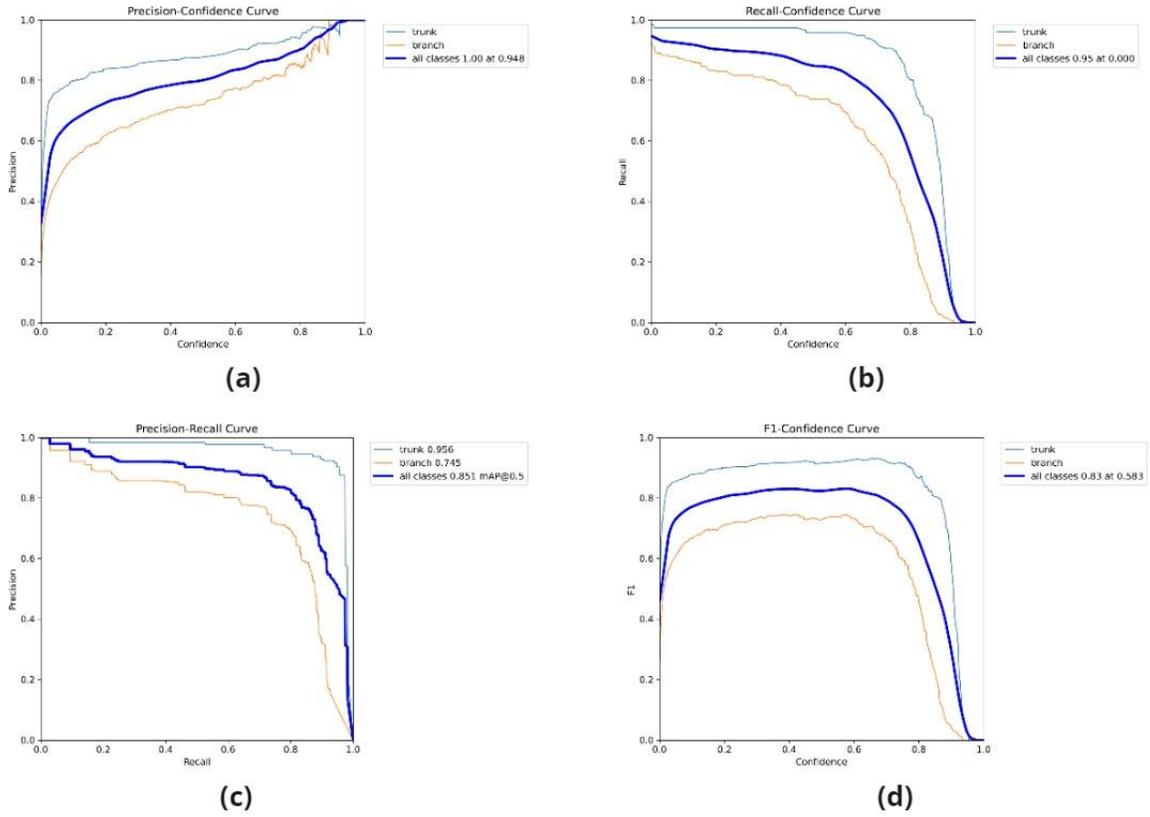

*Figure 9: Trunk and branch segmentation results achieved with YOLOv8 model; (a) Precision-confidence curve; (b) Recall-confidence curve; (c) Precision-Recall Curve; and (d) F1-Confidence curve*

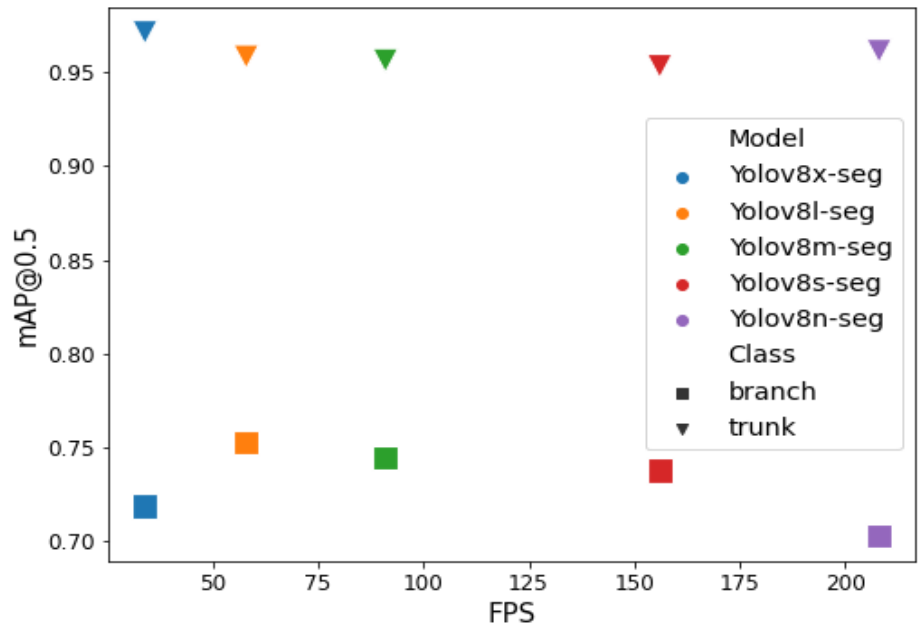

*Figure 10 : Model accuracy VS speed*

For branch detection and segmentation, the YOLOv8l-seg model outperformed all other models in terms of mAP and FPS, while YOLOv8n-seg had the lowest mAP and FPS values for branch detection and segmentation. However, it is noted that all models perform relatively worse for branch segmentation, with mAP around 0.75 or lower. When interpreting the mAP@0.5 vs FPS graph, it is important to consider the trade-off between accuracy and speed. In this case, the YOLOv8x-seg and YOLOv8l-seg models are the most accurate and fastest for trunk and branch detection and segmentation, respectively. However, it is also important to note that the branch segmentation task is comparatively challenging because of smaller dimension and complex structure in the image space, and relatively worse performance compared to the same with trunks was as expected.

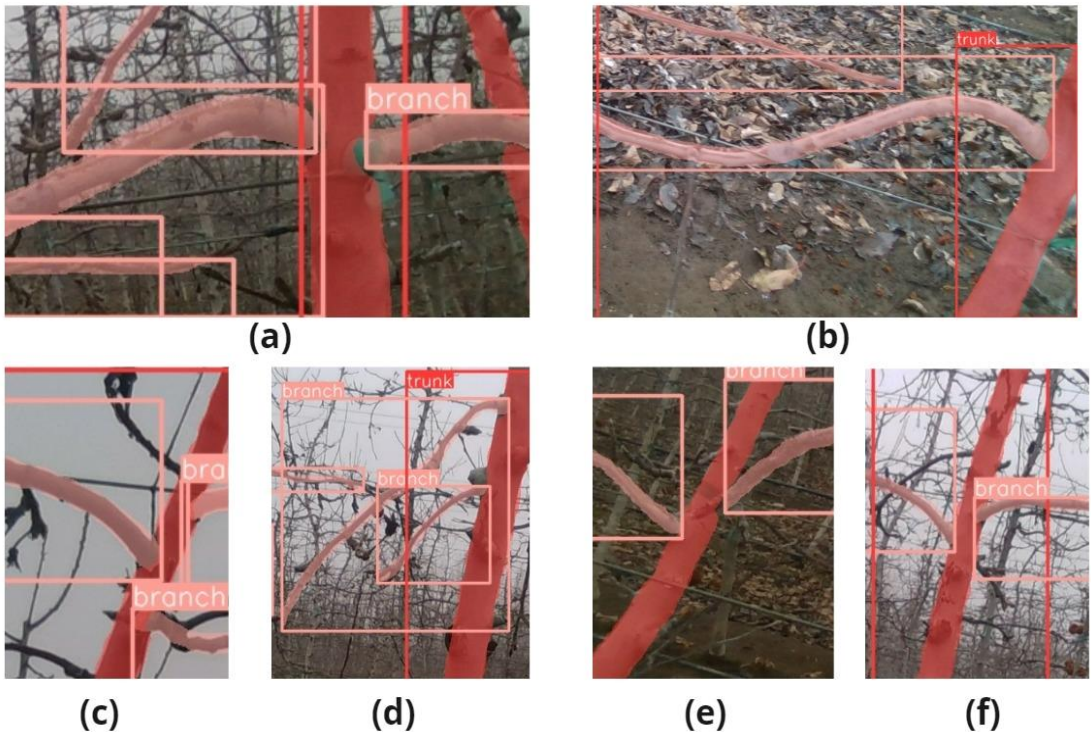

*Figure 11: Detection and segmentation of tree trunks and branches in variable lighting condition and variable locations; (a, d, and f) Robust detection and segmentation of trunks and branches in the presence of complex background and low light at different tree heights; (b) detection and segmentation with a noisy background in the lower part of tree; (c) segmentation in cloudy and low light condition and in top part of the tree; (e) segmentation in low light and middle part of the tree*

The results of this study suggest that the YOLOv8 family of models is effective for object detection and segmentation in an apple orchard setting. The mAP@0.5 vs FPS graph provided a useful tool for evaluating model performance, as it allowed for a trade-off between accuracy and speed to be considered. It is important, however, to carefully consider the specific use case and task requirements when selecting a model. While we have suggested the YOLOv8x-seg and YOLOv8l-seg models for trunk and branch detection and segmentation, respectively, it is worth noting that the use of two separate models may not be practical, and further research is needed to find an optimal model for the specific task at hand. Regarding the examples of detection and segmentation in Figures 11 and 12, it is important to discuss them separately. Figure 11 provides examples of trunk and branch detection and segmentation using the YOLOv8 algorithm, while Figure 12 presents some instance segmentation results. We can observe from Figure 12 that the model was not able to accurately identify branches in some cases, indicating a need for further improvement in branch segmentation performance.

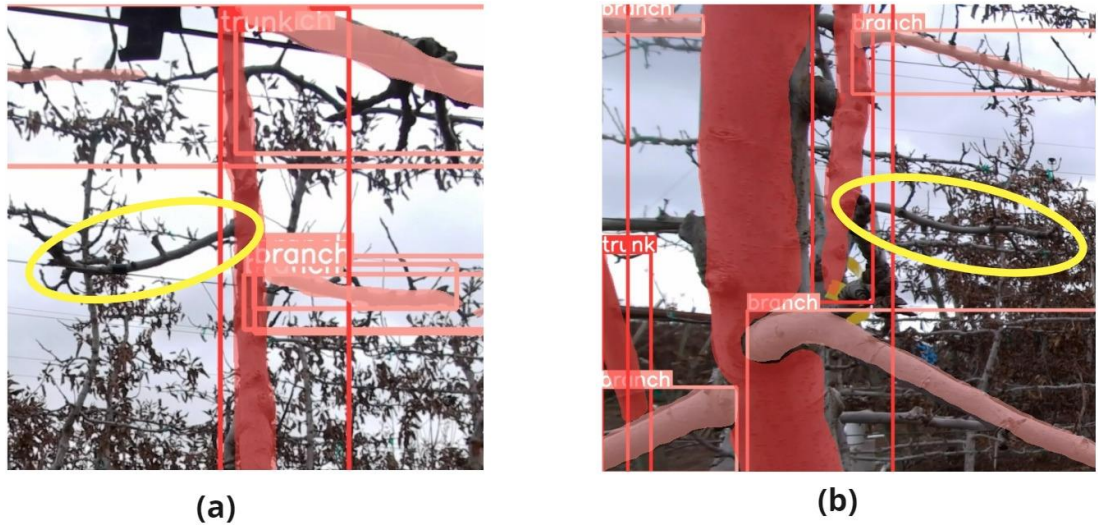

*Figure 12: Examples of unsuccessful detection of branches; (a) Caused by low light condition; and (b) Caused by complex branch sample due to low light and shadow condition*

In spite of the high accuracy achieved in trunk and branch detection and segmentation, the YOLOv8 model still generated some false positives and false negatives in some cases. For instance, Figure 12 (a) illustrates a case where the model failed to identify a branch segment, indicated by the yellow circle. This failure to detect the branch was most likely caused by limitations in the model training. Similarly, the circled region in Figure 12 (b) shows where YOLOv8 failed to identify the branch due to the limited number of samples used in this study for training. To address this issue, (Verma et al., 2018) suggested that training the model with a larger dataset, containing more input features, can significantly improve the model's generalization ability to new and unseen data. Moreover, a larger dataset can enable the model to capture the subtle differences in branch structures, such as those present in different tree species, ages, and environmental conditions during the dormant season. Additionally, a larger dataset can help to mitigate the risk of overfitting, where the model becomes too specialized to the training dataset and performs poorly on new samples.

Previous studies have utilized Mask R-CNN, a two-stage deep learning technique, for branch detection during the dormant season of apple orchards. However, the results achieved in our study show superior performance in terms of both detection accuracy and speed. Further optimization of the feature extraction process and exploration of more sophisticated machine learning algorithms can help to improve the model's accuracy. It is important to validate the model's predictions on new data to ensure generalization to unseen examples of branch position in natural orchard environments. To achieve this, a larger number of training samples are required.

### 3.2 Branch diameter estimation

The study showed robust capability in terms of estimating the diameter of apple tree branches. The approach of applying PCA on 3D point clouds of the segmented branch mask and drawing a perpendicular line (normal) to the point orientation was an effective method to estimate branch diameter in a natural orchard environment. To validate the diameter estimation method applied in this study, 43 samples of ground truth were compared with the predicted diameter value. Figure 13 shows the relationship between the actual value and predicted value of the branch diameters using the purposed machine vision system. The model achieved an RMSE of 2.08, which means that on average, the machine vision system's predictions for branch diameter are off by 2.08 mm compared to the ground truth measurements. Furthermore, a correlation coefficient of 0.82 was achieved which indicated a strong positive linear relationship between the predicted and actual diameter measurements. This means that as the predicted diameter measurements increase, the actual diameter measurements also tend to increase, and vice versa. This suggests that the study can generate more accurate results upon increasing the input dataset sample. The strength of the relationship suggests that the machine vision system's predictions for diameter estimation are generally reliable and consistent with the actual field measurements.

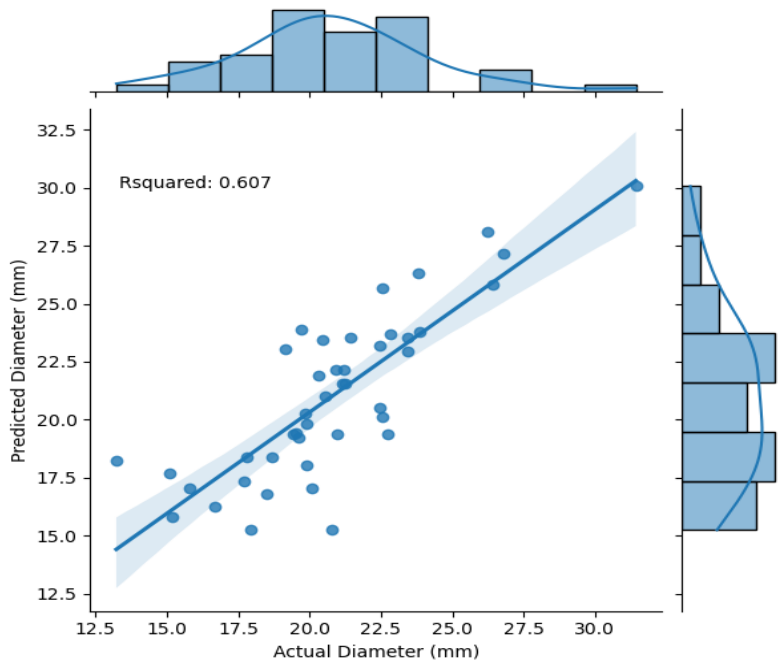

*Figure 13: Scatter plot for branch diameter estimation of 43 branch samples*

Likewise, over-masking and under-masking has been noticed during the segmentation of YOLOv8 model. When its over-masks the region, the estimated diameter is greater than the actual diameter since nearby secondary branch points were masked as primary branch points because of the over-masking. In the case of under-masking the region, the estimated diameter would be smaller than the actual diameter because the diameter calculated would only be for the masked area. The figure 14 (a) and (b) illustrates two scenarios where YOLOV8 model over-masked and under-masked the branch region respectively which have affected the diameter estimation accuracy at those regions.

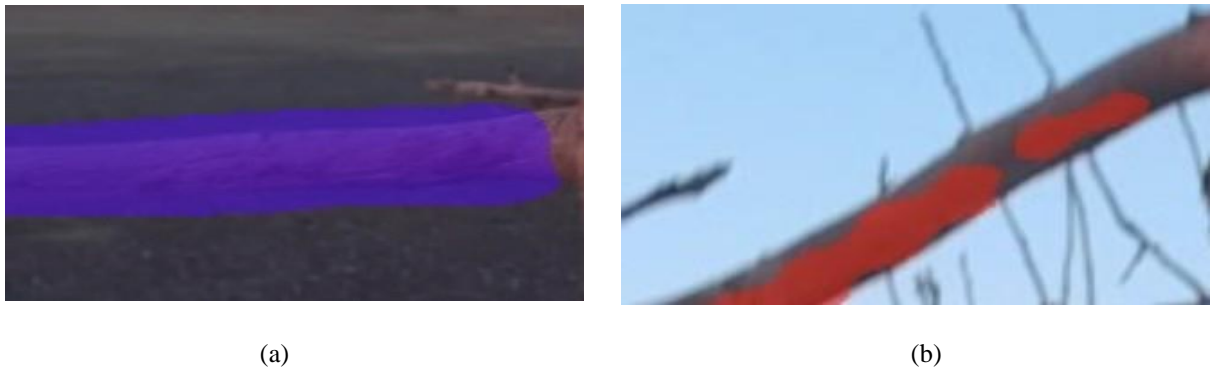

(a)                                                                                              (b)

*Figure 14: (a) over-masking of the branch segment  (b) under-masking of the branch segment*

After converting the diameter value into limb cross-sectional area (LCSA) by using the property of area of the circle $A = \pi d^2/4$, where A is the area and d is the branch diameter), estimated crop-load in each branch by the machine vision system was compared with the ground truth value. Figure 15 shows the normalized crop-load deviations for two category of diameters i.e 10-20 mm in diameter (smaller range) and the other one is 20-30 mm in diameter (bigger range), and Figure 16 shows the data pattern of 43 branch samples for target crop load estimation.

The purposed machine vision-based crop-load estimation system achieved RMSE of 3.95 indicating that on average, the estimated crop-load may differ from the actual crop-load by up to 3.95 units. In the context of desired crop-load estimation during dormant season application, this means that the machine vision system can estimate the number of crops with a certain degree of accuracy, but there may still be some degree of error in the estimation. Thus, the estimated crop-load should be used as a guide, and farmers should also rely on their expertise and experience to make appropriate management decisions. The mean absolute error (MAE) of 2.99 was achieved for the desired crop-load estimation which indicates that, on average, the estimated crop-load values produced by the machine vision system are off by 2.99 units from the true values. For example, if the true crop-load value of a particular plant is 10, the estimated value by the machine vision system may be anywhere between 7.01 and 12.99. This level of error is relatively low and suggests that the machine vision system is performing well for crop-load estimation.

In terms of application, crop-load estimation is a critical aspect of fruit and crop management. Accurate crop-load estimation helps farmers determine the optimal time for harvesting, manage crop-load and yield, and allocate resources effectively. With a machine vision system that can provide accurate crop-load estimates, farmers can save time and resources, increase efficiency, and improve overall crop quality and productivity. However, for the complex object such as branch in apple orchards, this value has significant value.

By using this approach, growers can optimize their yields by producing high-quality fruit while also maintaining the overall health and productivity of the tree. However, it's important to note that the "6 apples per cm limb cross-section area" guideline is just a rough estimate, and growers may need to adjust the number of apples per limb based on a variety of factors, including the variety of apple, the age and health of the tree, and the growing conditions. Recently, Ranadeep et al (Sidhu et al., 2022b) have described the effects of different crop-load and thinning methods on the yield, nutrient content, fruit quality, and physiological disorders in 'Scilate' apples by studying plant physiology under 3 crop-load management approaches 3, 6, and 12 fruits per cm² LCSA. The study found that 6 fruits per cm² LCSA was the most effective method of managing crop-load and optimizing the fruit quality.

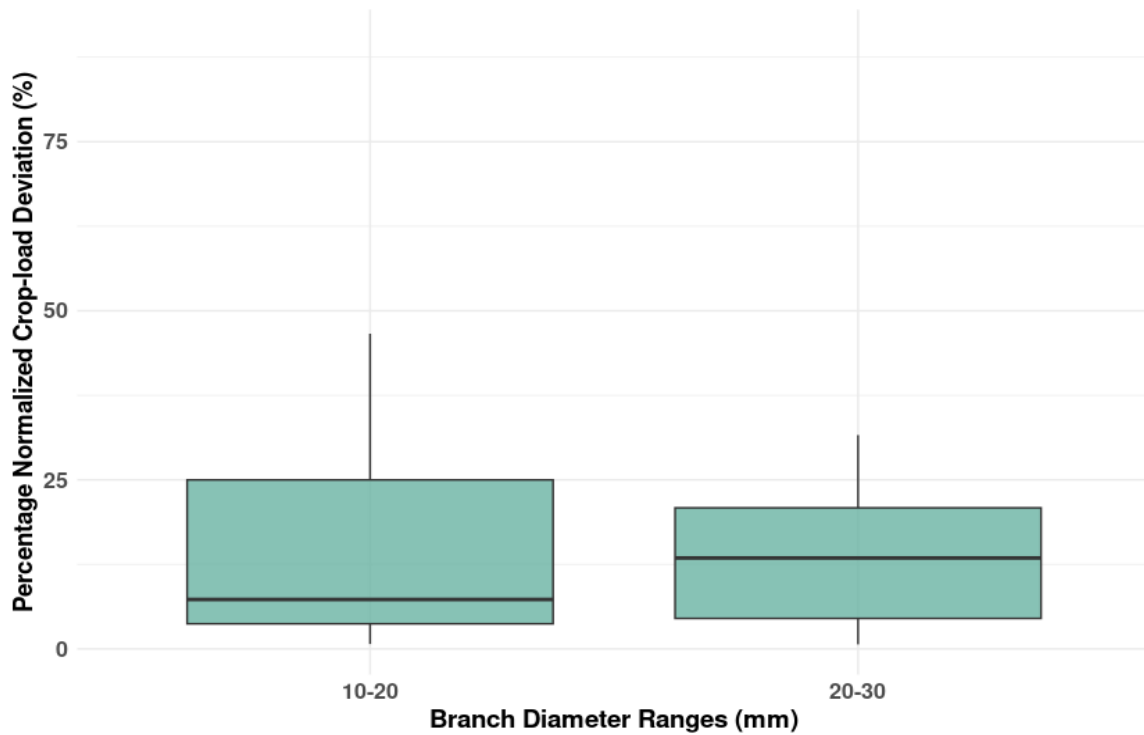

*Figure 15: Target crop load deviation plots (Normalized) for branch diameter estimation*

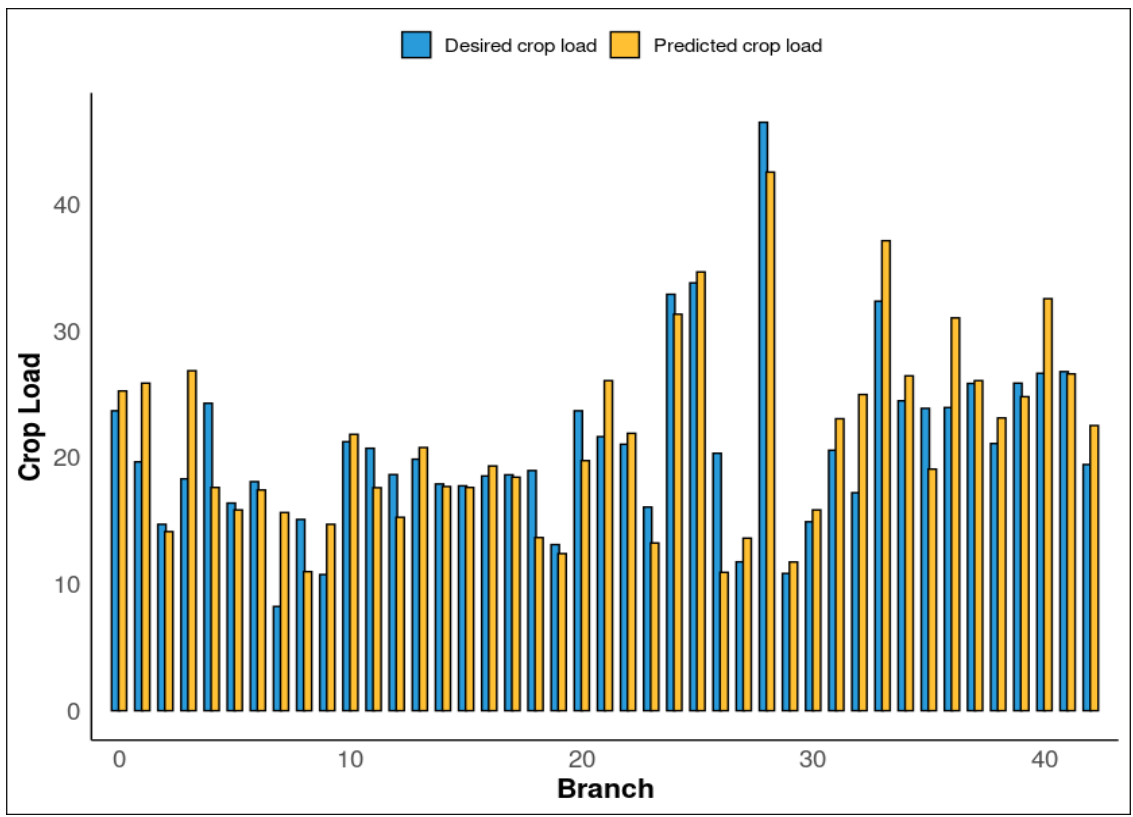

*Figure 16: Comparison of crop load estimation accuracy of the purposed machine vision system with ground truth*

    This kind of research is significant because it addresses a crucial need in orchard agriculture. Accurate estimation of crop-load is essential for optimal fruit yield, quality, and profitability in commercial orchards. Traditionally, crop-load estimation has been performed manually, which is time-consuming, labor-intensive, and prone to errors. The proposed machine vision system offers a more efficient and accurate alternative to manual crop-load estimation. By automating the process, the system can provide timely and accurate information about the crop-load, allowing farmers to adjust their management practices accordingly to optimize fruit yield and quality. Additionally, the system can reduce labor costs and improve productivity, making it a valuable tool for commercial orchard management. Therefore, this research is significant in advancing agricultural technology and improving the efficiency and profitability of commercial orchards.

    One technical benefit of this system is its ability to accurately estimate crop-load without causing any damage to the fruit or the tree. Traditional methods of crop-load estimation involve physically counting fruits, which can be labor-intensive and can cause damage to the fruit or the tree. The machine vision-based system eliminates the need for physical counting and provides non-invasive crop-load estimation. This can lead to improved fruit quality, reduced tree damage, and increased productivity.

    Another scientific benefit of the purposed machine vision system for estimating crop-load based on LCSA can also be helpful for automated pruning in orchards. During the dormant season, when the trees have shed their leaves, the machine vision system can accurately estimate the number of crops on each branch based on the LCSA, without the need for human intervention. This information can then be used to determine which branches need to be pruned to achieve the desired crop-load. Automated pruning systems that use machine vision can greatly increase efficiency and accuracy compared to manual pruning (Elfiky et al., 2015; You et al., 2022). By accurately estimating the desired crop-load on each branch, the machine vision system can optimize the pruning process to maximize yield and minimize waste. This can lead to increased profitability for orchard farmers and can also reduce the environmental impact of orchard agriculture by minimizing the use of chemicals and other inputs. Overall, the combination of

machine vision-based crop-load estimation and automated pruning has the potential to revolutionize the way orchard agriculture is practiced.

Additionally, the purposed machine vision system can also be helpful in green fruitlet thinning as it provides an indirect measure of the potential fruiting capacity of the branch. The number of fruiting positions on a branch is directly related to the branch's cross-sectional area. Therefore, by estimating the branch's LCSA using a machine vision system, it is possible to estimate the potential number of fruiting positions on that branch. This information can be used to optimize fruit set and yield by selectively removing excess fruitlets during the green fruitlet thinning process. By removing fruitlets from branches with a higher estimated LCSA, growers can ensure that the remaining fruitlets have sufficient resources to develop into high-quality fruits. This approach can help to improve fruit size, reduce biennial bearing, and increase overall orchard productivity.

Furthermore, the advantage of this kind of system is its ability to optimize the use of resources such as water, fertilizer, and labor. By estimating the optimal crop-load for each branch, the system can help farmers adjust their management practices accordingly, which can lead to improved resource efficiency and reduced costs. This can also help in reducing the environmental impact of orchard agriculture by minimizing the use of resources such as water and fertilizer. Additionally, the system can provide highly accurate information on the crop-load, allowing farmers to make timely and informed decisions about their management practices. Overall, the machine vision-based target crop-load estimation system can provide significant benefits to orchard agriculture by improving productivity, reducing costs, and minimizing environmental impact.

## 4. Conclusion

In this study, a machine vision system was developed for desired crop-load estimation in branches of commercial orchards. The performance of the system was evaluated using the YOLOv8 model for trunk and branch segmentation, achieving high precision, recall, and f1 scores for trunk segmentation up to 0.888, 0.974, and 0.922 and for branch segmentation up to 0.739, 0.789, and 0.742, respectively. The machine vision system also demonstrated good accuracy for the diameter estimation of branches and LCSA calculation with an RMSE of 2.08. The crop-load estimation was achieved using branch LCSA, which enabled the machine vision system to estimate the desired or target number of crops with an MAE value of 2.99 by employing the farm management approach of "6 fruits per cm square LCSA". Thus, the proposed machine vision system provides a promising solution for efficient and accurate crop-load estimation in commercial orchards. To improve the accuracy of the system, further research can focus on developing deep learning models with larger datasets and exploring different image processing techniques. Additionally, future work can investigate the feasibility of integrating the machine vision system with automated pruning systems to optimize orchard management practices. Furthermore, the use of other crop-load estimation methods, such as fruit count, can be explored to enhance the accuracy of the system. Overall, the application of machine vision technology in orchard agriculture can lead to more efficient and sustainable orchard management practices.

**Acknowledgment:** The research is funded by AgAID Institute - Agricultural AI for Transforming Workforce and Decision Support, and United States Department of Agriculture National Institute of Food and Agriculture (USDA NIFA). The authors gratefully acknowledge Prof. Dr. Matthew Whiting for the guidance and Dave Allan (Allan Bros., Inc.) for providing access to the orchard during the data collection and field evaluation.

**Author Contributions**
D.A and R.S conceptualized, designed and performed the investigation, data analysis and wrote the manuscript. M.C designed the integrated camera system and provided critical reviews. M.K provided critical reviews and edited the manuscript.
Note: All authors reviewed and approved the internet archive version of this study and is under submission to *computers and electronics in the agriculture* journal.